The Expressions of Depression and Anxiety in Chinese Psycho-counseling: Usage of First-person Singular Pronoun and Negative Emotional Words

Lizhi Ma[1], Tong Zhao[4], Shuai Zhang[2,3], Nirui Song [2], Hongliang He[2,3], Anqi Li[2,3], Ran Feng[3], Huachuan Qiu[2,3], Jingsong Ma[5], and Zhenzhong Lan[3]

[1] Jing Hengyi School of Education, Department of Psychology, Hangzhou Normal University

[2] Zhejiang University

[3] School of Engineering, Westlake University

[4] Queen Mary University of London

[5] School of Early Childhood Education, Hangzhou Polytechnic

**Author Note**

Lizhi Ma https://orcid.org/0000-0001-5528-8396, correspondence concerning this article should be addressed to Lizhi Ma, Jing Hengyi School of Education, Department of Psychology, Hangzhou Normal University, 311121, Hangzhou, Zhejiang, P.R.China; E-mail: malizhi@hznu.edu.cn. The current work belongs to part of "Xihu-xinlin" online psycho-counseling research.

**The Expressions of Depression and Anxiety in Chinese Psycho-counseling: Usage of First Person Singular Pronoun and Negative Emotional Words**

As early as the beginning of the 20th century, Sigmund Freud, the originator of the psycho-therapeutic approach to psychoanalysis, suggested that individuals' habits of speech, slips of the tongue, and other linguistic expressions to a certain extent represent people's subconscious hidden motives or concerns (Freud, 1989). In psycho-counselling, clients mainly talk to the counselor about their difficulties through linguistic expressions, and recent studies have also shown that individuals' verbal expressions often reflect their current psychological and cognitive states (Pennebaker et al., 2003). And an individual's depression or anxiety can be expressed through words, such as the publication of texts on social media (Bathina et al., 2021; Coppersmith et al., 2015; Qiu et al., 2012), thematic writing of emotion (Huang et al., 2019; Smirnova et al., 2018). For example, compare to the healthy control, individuals with depressive or anxious tendencies or symptoms use more the first person singular pronoun, negative emotion words, negatives, tentative words etc.(Miner et al., 2022; O'Dea et al., 2021; Qiu et al., 2012; Smirnova et al., 2018; Wei et al., 2021). However, most of such studies have been conducted in English contexts, and those in Chinese contexts have focused on unilateral text postings by users on social media platforms, rather than interactive dialogues between counselors and clients in the context of psycho-counselling (Cheng et al., 2017; Li et al., 2020).

In existing studies that have explored psycho-counselling dialogues, researchers quantified linguistic expressions in psycho-counsellings through the Linguistic Inquiry and Word Count Dictionary (LIWC) (Lin et al., 2020; Tausczik & Pennebaker, 2010) and analyzed their relationship with subjectively rated counselling effectiveness by the clients (Althoff et al., 2016). In the psycho-counselings, clients and counselors discussed the past and present in the first and middle stages of counselings, and explored the future in the later stage of counselings. Meanwhile, the clients usually used more negative emotional words in the first stage of counselings, and increased the use of positive emotional words in the later stage of counselings. Moreover, the clients who expressed more future-related and positive emotional vocabulary in the latter part of the counselings tended to perceived the psycho-counselling as more effective in soothing their emotional feelings compared to those who expressed less of this type of words. Notably, the depressed clients used more first-person pronouns and anxious clients used more quantifiers for their expressions compared to those who had a normal mental state (Miner et al., 2022; Wei et al., 2021). Thus, the clients' linguistic expressions in psycho-counselings can be quantified to a certain extent by means of psychologically meaningful vocabulary and are linked to their subjective feelings (Freud, 1989).

Although little research of Chinese psycho-counseling has systematically explored the characteristics of linguistic expressions in the counseling dialogues related to depression and anxiety states, and whether linguistic expressions change with the level of depression and anxiety of individuals (Malgaroli et al., 2023), it has been shown that an individual's expressions change as their physical and mental health state changes. Campbell and Pennebaker (2003) found that changes in an individual's writing style correlated with their level of physical fitness. Their experiment correlated the semantic similarity between participants' consecutive emotional writings or writings about traumatic events with their number of trips to the hospital for medical checkups. The results showed that the smaller the similarity in stylistic semantic space and personal pronoun semantic space between each of the participants' writings, the lower the number of trips they made to the hospital, indicating a higher degree of improvement in their health status. The phenomenon of an individual's writing style changing in response to individual's state of physical health also provides evidence that an individuals' physical and mental improvement is reflected in changes in the cognitive styles (Beck, 1979).

However, existing research is limited to the psycho-counseling in the Western English context or social media publishing in the Chinese context, and barely address the field of Chinese psycho-counselings. Moreover, specific expressions in the Chinese context, such as subject omission (Chen, 2009), and different attitudes toward emotions between Chinese and Westerners, such as the tendency of Easterners to express emotions in a calm manner while Westerners tend to express emotions in an agitated manner (Lim, 2016), may differentiate linguistic expressions in the Chinese context from the English context. Therefore, in order to explore the linguistic expressions affected by the degree of depression and anxiety in the Chinese psycho-counseling context, the current study focuses on comparing the differences in the use of first-person singular and negative emotional words in the Chinese psycho-counseling for clients with different degrees of depression and anxiety.

## Method

**Counseling Setting and Design**

An online platform called 'Xinling' was employed to collect all the psycho-counseling dialogues. The platform was developed by Deep Learning Lab of Westlake University and designed specially for the service of online text-based psycho-counselings. The data collected by the platform were stored on the Ali Cloud encrypted by Advanced Encryption Standard 256 algorithm (AliClound, 2023).

The design of the experimental setting included setting for both client and counselors. For the clients, they were encouraged to complete at least seven consecutive online psycho-counselling sessions, with no more than two weeks between sessions to ensure effectiveness (Kadera et al., 1996). They would receive participation fee after attending 10 consecutive online psycho-counselling sessions free of charge. At the same time, before the start of each counselling session, the clients should fill in the Self-rating Anxiety Scale (SAS, Zung, 1971) and Self-rating Depression Scale (SDS, Zung, 1965) to reflect their depression and anxiety status in the past week, meanwhile, they could also pay for the counseling if they wished to continue after the 10 sessions. For the counselors, they were also informed that they would give at least 7 online text-based psycho-counselings to each client and the clients would fill SAS and SDS at the start of counselings. The counselors were also required to provide the online text-based counselings as professional as face-to-face counselings. For both clients and counselors, they were informed that they could cease the counselling or require transfer anytime after the counselings started. The research project was granted ethics approval by the Westlake University Research Ethics Committee (20220519LZZ001).

**Participants Recruitment**

*Client*

All the participants were recruited via advertising, word-of-mouth and snow-balling. Considering the limitations of online text-based counseling in terms of dealing with mental disorders and crisis situations (Haque & Rubya, 2022), the recruitment contained two stages to ensure all the participants were suitable for the online text-based counseling. At the first stage, all the individuals signed up for the study were required to complete the self-report symptom inventory (SCL-90) (Derogatis et al., 1973; Wang et al., 1999) to confirm that the participants were not suffering from severe psychological problems and psychiatric symptoms. As a result, 163 individuals signed up but eight of them failed to complete the checklist. Another seven of them were excluded due to their T scores of depression, anxiety or psychiatric symptoms were above 2.5, indicating mild to severe level of psychological problems (Wang et al., 1999). Meanwhile, the seven individuals were suggested to visit the clinic for a professional check.

At the second stage, researchers with psychological background contacted the rest 148 individuals one by one via phone call to further examine their mental health condition and motivation of participation; meanwhile, the researchers also introduced the procedure of the study and acquire consents from the individuals who decided to participate. Consequently, 46 individuals were excluded from the study (21 were only curious about the study and did not have any issues to discuss with psycho-counselors, 13 were unable to participate due to time clash, 10 were having ongoing psycho-counselings or medical treatments, 2 lost of contact). Thus, 102 participants were recruited to attend the online counselling after the two stages. Although 20 of them dropped out as the counselings proceeded (19 due to time clash, 1 transfer), 82 adult participants completed the entire study (55 females, 19- to 54-year-olds, $M = 27.62$ years, $SD = 5.94$) and received 300 Chinese Yuan for participation.

*Counsellor*

We invited nine qualified counselors[1] to participate the current research (7 females, 25- to 45-year-olds, $M = 34.67$ years, $SD = 7.45$). Three of them were experienced psycho-counselors with over 10 years' and 4000 hours' counseling experience (professional), another three were senior counselors with over 5 years' and at least 400 hours' experience (senior), the rest three were junior counselors with counseling experience less than 5 years and 300 hours (junior). Each psycho-counselor received reasonable participation fee based on the level of experience and the amount of counselings they completed. All the clients and counselors gave their consent to participate in the current research, use the data collected for non-profit academic research and received participation fee.

---

[1] Seven counselors hold the national practitioners' certificates of psycho-counselling or psychotherapy; two junior counsellor had completed professional counselling training for at least 200 hours and accomplished more than 100 hours' counselings independently under supervision before participated the current research.

In the end, we collected 859 counselings in total of which 735 counselings with depression ($M = 49.52$, $SD = 12.83$) and anxiety ($M = 44.88$, $SD = 8.21$) scores reported (Table 1). The following analyses were based on the 735 counselings.

**Quantify First-person Singular Pronoun and Negative Emotional Words**

Clients' expressions of first person pronoun and negative emotional words were quantified by LIWC-22 software (Boyd et al., 2022) using the simplified Chinese LIWC dictionary (Lin et al., 2020) based on the frequency (e.g., the number of fist-person pronouns/the entire words clients typed in one counseling session).

Table 1

The Descriptive Information of the Collected Counsellings

| Role | Number | Utterances | Utterances per session Mean (min-max) | SD | Length per utterance Mean (min-max) | SD |
|---|---|---|---|---|---|---|
| *The Descriptive Information of the Collected 859 Counsellings* | | | | | | |
| Counsellors | 9 | 32329 | 37.64 (7-141) | 13.51 | 14.32 (1-35) | 4.72 |
| Clients | 82 | 32159 | 37.44 (8-142) | 13.52 | 17.76 (1-84) | 9.75 |
| Total | 91 | 64488 | 75.07 (15-283) | 27.02 | 16.03 (1-84) | 5.81 |
| *The Descriptive Information of the 735 Counsellings with Depression and Anxiety Scores* | | | | | | |
| Counsellors | 9 | 27744 | 37.75 (7-89) | 12.61 | 14.31 (1 - 35) | 4.68 |
| Clients | 82 | 27595 | 37.54 (8-89) | 12.61 | 18.23 (1 - 84) | 10.03 |
| Total | 91 | 55339 | 75.29 (15-178) | 25.2 | 16.26 (1 - 84) | 5.92 |

## Result

We employed general linear mixed effect model (GLMEM) to exam the relations between clients' usage of first-person singular pronoun, negative emotional words and their degree of depression and anxiety. Data were analyzed in RStudio (version 2023.06.0; Allaire, 2012). We used the lmer function from the lme4 package to fit linear mixed- effect models (LMEMs) in R (Douglas Bates et al., 2015). Generally speaking, we fitted the different models with clients' usage frequency of first-person singular pronoun or negative emotional words as dependent variable, clients' depression or anxiety scores as independent variable, and by-clients, by-counselors random intercepts. In the result section, first, we reported the result of whether clients' usage of first-person singular pronoun and negative emotional words change with depression degree, then those change with anxiety degree.

### The Relation between First-person Singular pronoun and Depression

The result of GLMEM with the frequency of first-person singular pronoun as dependent variable shows that the usage frequency of first-person singular pronoun ($M = 5.31$, $SD = 2.27$) has no relation with depression scores ($\chi^2(1) = 1.16$, $p = .28$), indicating the clients used similar amount of first-person pronoun no matter how depressive they were (see Figure 1).

Figure 1

*The Trend of First-person Singular Pronoun and Negative Emotional Words Based on Depression Score*

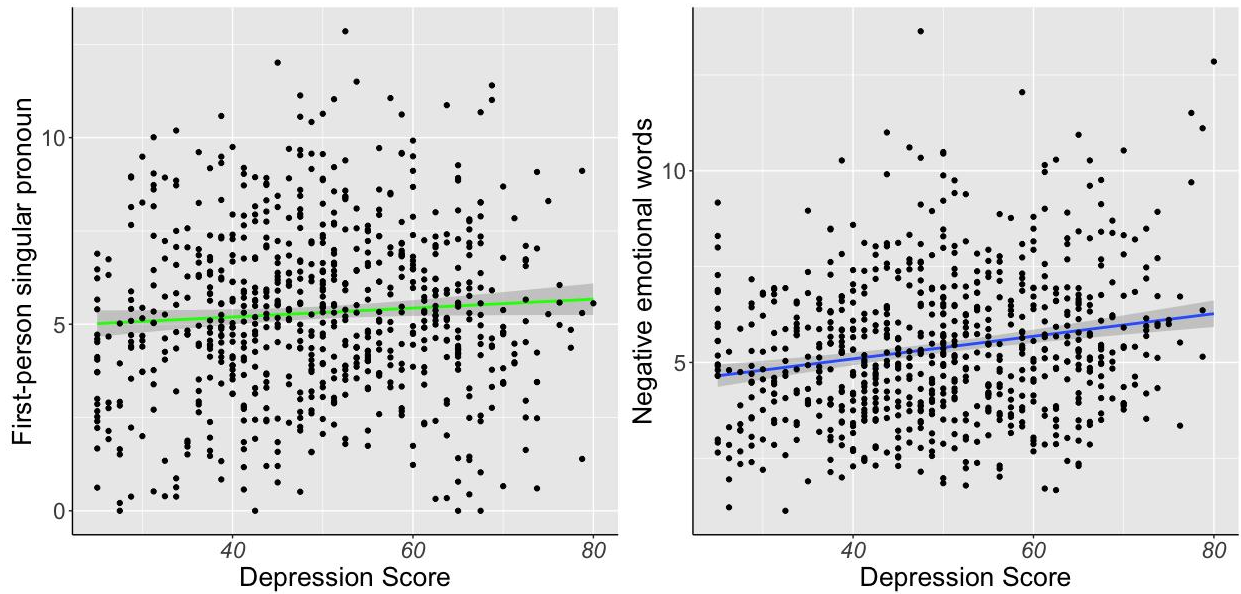

*Note.* The black dots represent the individual data points of each clients in each counseling sessions.

**The Relation between Negative Emotional Words and Depression**

The model with the frequency of negative emotional words as dependent variable shows that the usage of negative emotional words (*M* = 5.37 , *SD* = 1.89) relates to clients' depression score positively ($\chi^2(1)$ = 15.77, *p* < .001), indicating the more depressive the clients were, the more negative emotional words they expressed in the psycho-counseling (see Figure 1).

**The Relation between First-person Singular Pronoun and Anxiety**

Similarly, the model with the frequency of first-person singular pronoun as dependent variable shows no significant relation between the frequency of first-person singular pronoun and clients' anxiety score ($\chi^2(1) = 0.15$, $p = .69$), indicating the clients used similar amount of first-person singular pronoun regardless of how anxious they were in the Chinese psycho-counseling (see Figure 2).

**The Relation between Negative Emotional Words and Anxiety**

The model with the frequency of negative emotional words as dependent variable shows that the usage of negative emotional words relates to clients' anxiety score positively ($\chi^2(1) = 8.66$, $p = .003$), indicating the more anxious the clients were, the more negative emotional words they expressed in the psycho-counseling (see Figure 2).

Figure 2

*The Trend of First-person Singular Pronoun and Negative Emotional Words Based on Anxiety Score*

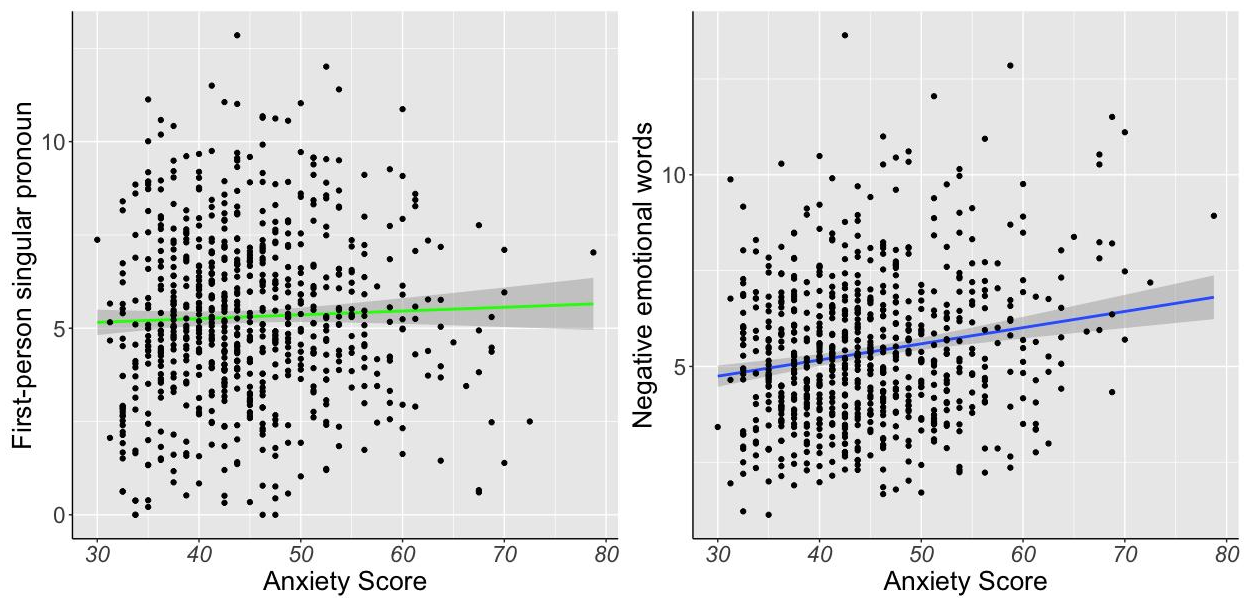

*Note.* The black dots represent the individual data points of each clients in each counseling sessions.

## Discussion

The current work analyzed whether the use of the first-person singular and negative emotional words differed between clients in different states of depression and anxiety in a Chinese online counseling conversation setting, and found that the more severe the level of depression and anxiety, the more emotional negatively the clients viewed and described things, but that the use of the first-person singular was not affected by the level of depression and anxiety of the clients. We discussed the former finding from the perspectives of therapeutic encouragement to express emotions during psycho-counseling and human nature to express negative emotions, and the later findings from the aspects of cultural differences between western and eastern world as well as conversational dynamics in psycho-counseling.

### Therapeutic Encouragement and Human Nature to Express Negative Emotions

On one hand, counselors and psychotherapists often encourage clients to articulate their feelings as part of the healing process. This encouragement is grounded in the belief that verbalizing emotions can help clients process their experiences(Jacobson, 1994), gain insight into their emotional states (Greenberg et al., 2008), and develop coping strategies(Trappler & Newville, 2007). In the context of depression and anxiety, where individuals may experience intense and overwhelming negative emotions, counselors might specifically prompt clients to describe these feelings(Ehrhardt et al., 1989), facilitating clients to confront and manage their emotions more effectively. On the other hand, from a humanistic perspective, expressing negative emotions can serve as a form of emotional catharsis, releasing pent-up emotional tension(Kashdan et al., 2007; Lieberman & Goldstein, 2006). For individuals experiencing depression and anxiety, articulating these feelings can provide a sense of relief and reduce the burden of carrying these emotions internally, for instance, expressing anger facilitates the improvement of life quality and decreases the feelings of depression (Kashdan et al., 2007). Meanwhile, expressing negative emotions is a way to seek connection and support. When individuals verbalize their distress, they are not only processing their own emotions (Greenberg, 2014) but also signaling their need for understanding and empathy (Graham et al., 2008). The act of expression itself, especially in a supportive and understanding environment, as in the psycho-counselings, can be therapeutic.

**Cultural Differences, Conversational Dynamics in Psycho-counseling and First-person singular Pronoun**

　　　A meta-analysis review (Oyserman et al., 2002) reported that Western cultures often emphasize individualism, where personal identity and expressions are strongly tied to individual experiences. This can reflect in language use, where first-person singular pronouns become a tool to assert individuality, especially in contexts discussing personal emotional states. Eastern cultures, including Chinese society, are more collectivist, focusing on the group and social harmony. Even in expressing personal distress, the linguistic style may prioritize a more collective or indirect approach, minimizing the focus on the self even when discussing personal issues. In terms of conversational dynamics in psycho-counseling, the psycho-counseling involves a directed, purposeful conversation between a client and a therapist, where the focus is on the client's thoughts, feelings, and experiences (Beck, 2020). Social media postings, however, are more about broadcasting to an audience, seeking validation, attention, or sharing information (Lyons et al., 2018). This difference in interaction nature could explain why the first-person singular usage doesn't increase with depression and anxiety severity in counseling settings. The conversational dynamic in counseling already centers around the client, making the excessive use of "I" unnecessary to emphasize their personal experience or to seek attention.

The present study elucidates that within the ambit of Chinese psycho-counseling, clients manifested an increase in the expression of negative emotions concurrent with heightened levels of depression and anxiety. Contrary to prior research, which posited an augmented utilization of first-person singular pronouns as indicative of depressive and anxious states in the context of English-speaking counseling sessions and social media interactions, such linguistic patterns do not appear to serve as reliable markers of depression or anxiety within the Chinese psycho-counseling framework. These findings underscore a distinct communicative function of social media VS psycho-counseling and may further reflect the underlying cultural dichotomies between the individualism prevalent in Western societies and the collectivism characteristic of Chinese culture.